\def\CLASSINPUTinnersidemargin{0.75in}
\def\CLASSINPUToutersidemargin{0.75in}
\def\CLASSINPUTbottomtextmargin{0.75in}
\documentclass[conference]{IEEEtran}
\IEEEoverridecommandlockouts
\usepackage{cite}
\usepackage{amsmath,amssymb,amsfonts}
\usepackage{algorithmic}
\usepackage{times}
\usepackage{graphicx}
\usepackage{multicol}
\usepackage{textcomp}
\usepackage{url}
\usepackage[bookmarks=true]{hyperref}
\hypersetup{hidelinks}
\usepackage[justification=justified]{caption}
\usepackage{xcolor}
\usepackage{multirow}
\usepackage{capt-of}
\usepackage[x11names,dvipsnames]{xcolor}
\usepackage{tabularx}
\newcolumntype{Y}{>{\centering\arraybackslash}X}
\usepackage{array}
\usepackage{pifont}
\usepackage{makecell}

\newcommand{\cmark}{\ding{51}}
\newcommand{\xmark}{\ding{55}}
\newcommand{\pmark}{$\sim$}

\def\BibTeX{{\rm B\kern-.05em{\sc i\kern-.025em b}\kern-.08em
    T\kern-.1667em\lower.7ex\hbox{E}\kern-.125emX}}
\usepackage{booktabs}
\begin{document}
\title{\vspace{0.25in}\textsc{\textbf{U}niversal \textbf{N}avigation \textbf{I}nterface}: Robot-Free Data for Wheeled Robot Navigation}

\author{ 
    Sarvesh Prajapati$^{1}$,
    Ananya Trivedi$^{1,*}$,
    Lorena Maria Genua$^{1,*}$,
    Drake Moore$^{1}$,\\ 
    Bruce Maxwell$^{2}$, and
    Ta\c{s}k{\i}n Pad{\i}r$^{1,\ddagger}$\\ \vspace{0.5em}\normalsize\textcolor{magenta}{\url{https://universal-nav.github.io/}}\vspace{-1em}
    \thanks{$^{1}$Institute for Experiential Robotics, Northeastern    University, Boston, MA.
    
    $^{2}$Khoury College of Computer Science, Northeastern University, Seattle, WA, USA.
    
    {Corresponding author: \tt\small prajapati.s@northeastern.edu}}
    \thanks{$^\ddagger$Ta\c{s}k{\i}n Pad{\i}r holds concurrent appointments as a Professor of Electrical and Computer Engineering at Northeastern University and as an Amazon Scholar. This paper describes work performed at Northeastern University and is not associated with Amazon.}
    \thanks{This research was funded, in part, by the Advanced Research Projects Agency for Health (ARPA-H) Agreement No. 140D042590012. The views and conclusions contained in this document are those of the authors and should not be interpreted as representing the official policies, either expressed or implied, of the U.S. Government.}%
    \thanks{$^*$ Denotes equal contribution.}
}

\makeatletter
\let\@oldmaketitle\@maketitle
\renewcommand{\@maketitle}{\@oldmaketitle
    \begin{center}
        \includegraphics[width=\linewidth]{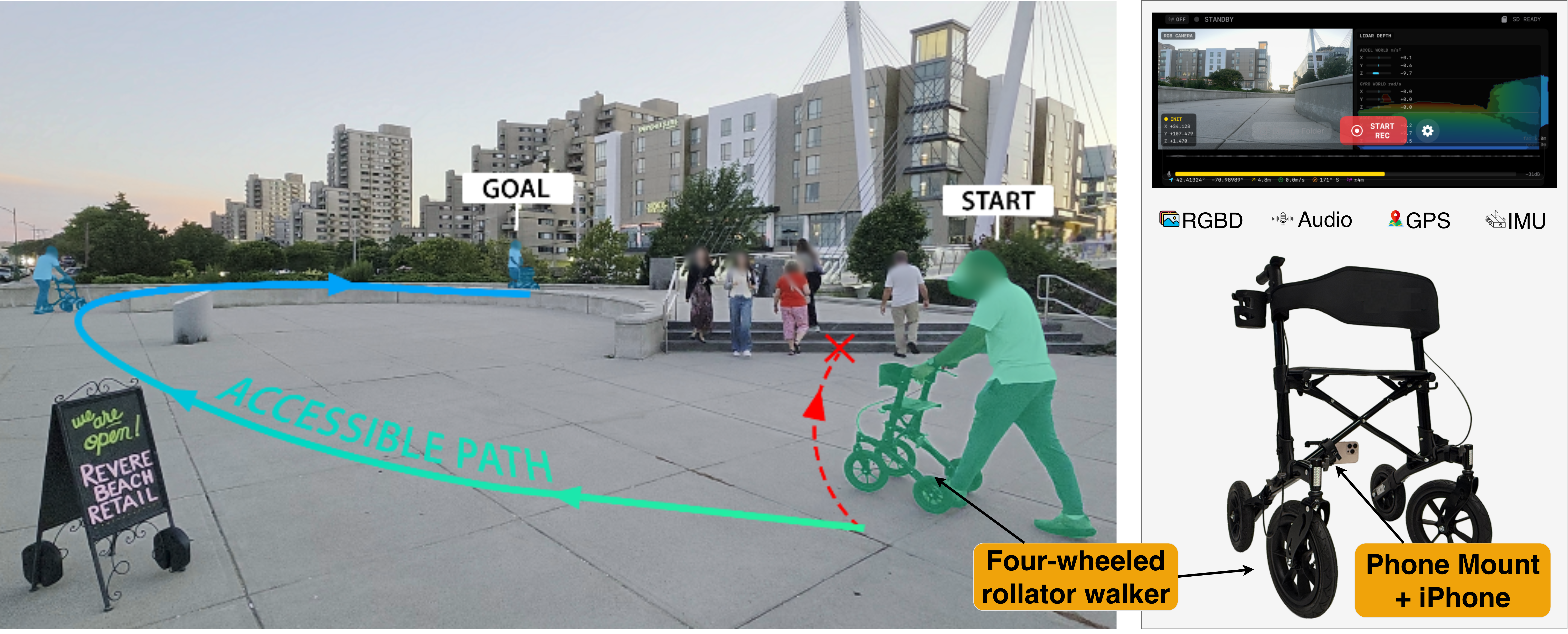}
        \setcounter{figure}{0}
        \vspace{-1em}
        \captionof{figure}{The \textsc{Universal Navigation Interface} (UNI) collects robot free wheeled-feasible navigation data without a robot, using a rollator and smartphone. Left: The rollator physically biases operators toward accessible paths, like ramps, instead of stairs. Right: The smartphone records synchronized multimodal data (RGB, depth, IMU, audio, GPS).}
        \label{fig:teaser}
    \end{center}
    \vspace{-1.5em}
}
\makeatother

\maketitle
\begin{abstract}
Collecting real-world navigation data for mobile robots typically requires platform-specific teleoperation, making large-scale collection expensive and difficult to scale. We introduce \textsc{Universal Navigation Interface} (UNI), a robot-free data collection paradigm that uses a four-wheeled rollator walker (rollator) and smartphone to collect physically constrained human demonstrations. Because the rollator cannot climb stairs, negotiate uncut curbs, or pass through narrow gaps, demonstrations are naturally biased toward wheeled-feasible routes. Using UNI, we collect 37.2\,km of real-world navigation data and recover metric trajectories that directly supervise goal-conditioned navigation models. Fine-tuning visual-navigation models on UNI reduces trajectory prediction error by 17.4--24.8\% on held-out UNI demonstrations. Evaluation on other navigation datasets shows benefits that vary by dataset and metric. We further demonstrate closed-loop transfer to a powered wheelchair in curb, staircase, and curb-cut scenarios. These results support low-cost physical proxies as a practical source of navigation supervision collected without the target robot.
\end{abstract}

\IEEEpeerreviewmaketitle
\setcounter{figure}{1}

\begin{table*}[t]
\centering
\caption{Representative datasets for visual navigation. \emph{Wheeled} denotes a physical wheeled
constraint during collection, rather than instructions or post-hoc filtering; \pmark\ denotes a corpus only partly collected under such a constraint. \emph{Robot-free} denotes that no robotic platform is
operated, and \emph{off-the-shelf}, consumer hardware with no purpose-built sensing rig.}
\label{tab:dataset_comparison}
 
\setlength{\tabcolsep}{6pt}
\renewcommand{\arraystretch}{1.25}
\footnotesize
\begin{tabularx}{\textwidth}{@{} >{\centering\arraybackslash}m{0.13\textwidth}
  >{\centering\arraybackslash}m{0.17\textwidth}
  Y Y Y c
  >{\centering\arraybackslash}m{0.20\textwidth} @{}}
\toprule
Dataset &
Collection platform &
Wheeled &
Robot-free &
Off-the-shelf &
Stationary\ (\%) &
Trajectory supervision \\
\midrule
\addlinespace[3pt]
 
SCAND~\cite{scand} &
\makecell{Teleoperated robots\\(wheeled + legged)} &
\pmark & \xmark & \xmark &
0.25 & Wheel odometry \\
\addlinespace[5pt]
 
\mbox{FrodoBots-2K}~\cite{frodo} &
\makecell{Teleoperated\\sidewalk robots} &
\cmark & \xmark & \xmark &
NR & GPS; wheel telemetry \\
\addlinespace[5pt]
 
EgoWalk~\cite{egowalk} &
\makecell{Human-worn\\stereo rig} &
\xmark & \cmark & \xmark &
3.10 & Stereo--inertial odometry \\
\addlinespace[5pt]
 
CityWalker~\cite{liu2025citywalker} &
\makecell{Web walking +\\driving video} &
\pmark & \cmark & \cmark &
NR & Normalized monocular VO \\
\addlinespace[5pt]
 
\mbox{Tartu/Milrem}~\cite{adl_milrem_2023} &
\makecell{Human-pushed\\golf trolley} &
\cmark & \cmark & \xmark &
filtered$^{a}$ & Visual odometry; GPS \\
\addlinespace[5pt]
 
\midrule
\addlinespace[3pt]
 
\textbf{UNI (ours)} &
\textbf{\makecell{Rollator +\\smartphone}} &
\cmark & \cmark & \cmark &
\textbf{8.14} & \textbf{\makecell{Depth-anchored metric\\trajectories}} \\
 
\bottomrule
\end{tabularx}
 
\vspace{2pt}
\begin{minipage}{\textwidth}
\footnotesize
$^{a}$Motion below 0.05\,m/s is removed during preprocessing.
\end{minipage}
\end{table*}

\section{Introduction}

Wheeled robots are increasingly being deployed on public sidewalks for last-mile delivery, inspection, and powered mobility. Learning-based navigation in these settings requires diverse, in-domain trajectory demonstrations, which are costly to collect. Autonomous driving has benefited from large-scale data collection and datasets that support perception and navigation research~\cite{waymoe2e,kitti}. Scaling data collection for sidewalk robots, however, requires deploying and supervising robotic platforms across varied pedestrian environments. For academic groups, these requirements limit the number of locations and conditions that can be covered; data collected by commercial fleets may remain proprietary. An accessible collection interface that operates without the target robot could therefore broaden participation and expand the diversity of navigation demonstrations.

Existing navigation data sources offer different tradeoffs. Robot-collected datasets~\cite{recon,sacson,gostanford,scand} provide demonstrations recorded directly on the robot, but require operating a robot. Human-worn rigs~\cite{nguyen2023musohu} and web video~\cite{liu2025citywalker} reduce this requirement, yet pedestrian routes may include stairs, uncut curbs, or narrow gaps that a wheeled robot cannot traverse. Filtering these routes afterward cannot recover the wheeled-feasible alternatives that were never demonstrated.

Manipulation has addressed a similar challenge by collecting demonstrations without the robot. The Universal Manipulation Interface (UMI)~\cite{umi} uses a hand-held gripper to collect demonstrations without a robot present while preserving task-relevant physical constraints. This motivates a corresponding question for navigation: can a simple physical proxy enable robot-free collection while biasing human demonstrations toward wheeled-feasible routes?

To address this, we introduce the Universal Navigation Interface (UNI) (Fig.~\ref{fig:teaser}): a commercial four-wheeled rollator walker (rollator) equipped with a smartphone. The collection hardware costs roughly \$250, excluding the smartphone, and requires no custom mechanical assembly. During collection, the operator pushes the rollator with all four wheels on the ground, favoring ramps and curb cuts over stairs and uncut curbs, while its width constrains passage through narrow gaps. These physical constraints bias demonstrations toward wheeled-feasible routes. A LiDAR-equipped smartphone provides RGB, depth, inertial, and GPS measurements in one device, avoiding the need to integrate separate sensors. We process these recordings offline into observation--goal pairs and metric future trajectories for training existing navigation models, without collecting demonstrations on the target robot.

Using UNI, we collected 37.2\,km of urban navigation demonstrations across 87 sessions in three cities. Fine-tuning visual navigation models like GNM~\cite{shah2023gnm}, ViNT~\cite{shah2023vint}, and NoMaD~\cite{sridharNoMaDGoalMasked2023} on UNI reduces trajectory error on held-out UNI demonstrations by 17.4--24.8\%. The recordings also include pauses during navigation, such as waiting at crosswalks and yielding to pedestrians. For ViNT, fine-tuning reduces stationary trajectory error by 65.7\%. We further demonstrate transfer from the rollator to a powered wheelchair through closed-loop navigation experiments.

To summarize, the contributions of this paper are:
\begin{itemize}
    \item \textbf{A low-cost instrument for robot-free navigation data collection.} We introduce UNI, which uses a rollator to physically constrain human demonstrations toward wheeled-feasible routes, decoupling navigation data collection from the target robot.

    \item \textbf{Metric real-world navigation supervision.} Using UNI, we collect 37.2\,km of multimodal navigation data and recover metric trajectories that can directly supervise existing goal-conditioned navigation models.

    \item \textbf{Navigation learning and embodiment transfer.} We demonstrate improved trajectory prediction across three navigation models and evaluate closed-loop transfer from the collection rollator to a powered wheelchair.
\end{itemize}

\section{Related Work}

A large body of prior work has focused on scaling data collection for robot navigation. Existing approaches span teleoperated robot datasets, human-collected navigation demonstrations, and supervision recovered from web-scale video, each trading off collection cost, embodiment fidelity, and scale. We organize prior work along these axes and discuss how UNI differs by decoupling data collection from the target robot while retaining a physical wheeled constraint during collection. Table~\ref{tab:dataset_comparison} summarizes how UNI compares with representative navigation datasets along these dimensions.

\paragraph{Robot-native navigation data} Most learning-based mobile navigation systems are trained from data collected directly on robotic platforms. Datasets such as RECON~\cite{recon}, SACSoN~\cite{sacson}, and SCAND~\cite{scand} pair observations with motion executed by robots, providing demonstrations that reflect the capabilities and physical constraints of the collecting platform. Models such as ViNT~\cite{shah2023vint} combine data from multiple robots and environments to learn more general navigation policies. While pooling existing datasets broadens training coverage, collecting new experience across locations and conditions still requires transporting, maintaining, and operating robotic platforms. UNI reduces this collection overhead by using a human-pushed rollator and smartphone to gather demonstrations without the target robot.

\paragraph{Human-collected and web-video navigation data}Human-worn datasets such as MuSoHu~\cite{nguyen2023musohu} and EgoWalk~\cite{egowalk} capture natural navigation and social interactions without robot teleoperation. These recordings provide useful supervision, with EgoWalk reporting improved robot navigation after fine-tuning on human demonstrations. Transfer nevertheless requires accounting for differences in human motion, sensor viewpoint, and robot capabilities. MuSoHu identifies gait-induced motion and viewpoint mismatch, while EgoWalk acknowledges that human demonstrations may include robot-infeasible maneuvers despite explicit collection guidelines.

CityWalker~\cite{liu2025citywalker} further expands coverage through large-scale online walking and driving videos, using visual odometry to generate action supervision without manual labeling. Its pipeline mitigates accumulated drift through short temporal windows and handles scale ambiguity by normalizing trajectories, rather than recovering their original metric scale. UNI complements these approaches by physically biasing collection toward wheeled-feasible routes and using recorded depth to anchor trajectory supervision in meters.

\paragraph{Robot-free demonstration collection} The manipulation community has demonstrated that useful robot supervision need not be collected with the target robot itself. The Universal Manipulation Interface (UMI)~\cite{umi} introduced a hand-held proxy gripper that preserves task-relevant sensing and action constraints while enabling portable, in-the-wild demonstration collection. This paradigm has since been extended across a broad range of manipulation settings, including aerial manipulation~\cite{umi-on-air}, and contact-rich or tactile tasks~\cite{tacumi}, demonstrating the value of appropriately designed proxy embodiments for scalable robot learning. UNI builds on the same principle for navigation: rather than reproducing the target robot's complete sensing and actuation stack, we preserve a physical constraint central to wheeled mobility---the demonstrated route itself must be traversable by a wheeled carrier.

The closest prior work to this idea in navigation is the Tartu/Milrem off-road navigation dataset~\cite{adl_milrem_2023}, which also collects demonstrations using a human-pushed golf trolley. Its collection platform, however, is a purpose-built sensing rig equipped with a ZED~2i stereo camera, an Xsens GNSS/INS unit, and three GoPro cameras, and was developed primarily for large-scale off-road navigation with geographic-map supervision. UNI instead asks whether the collection interface can be reduced to off-the-shelf consumer hardware while retaining the physical benefit of a wheeled proxy. A commercial rollator and smartphone provide a lightweight collection system for pedestrian-scale urban environments, while the rollator directly constrains demonstrations around infrastructure such as stairs, uncut curbs, curb cuts, ramps, and narrow passages. UNI additionally retains naturally occurring low-motion behavior such as waiting and yielding, rather than filtering it from the training corpus, and recovers depth-anchored metric trajectories for direct use by existing goal-conditioned navigation models.

\begin{figure*}
    \centering
    \includegraphics[width=\linewidth]{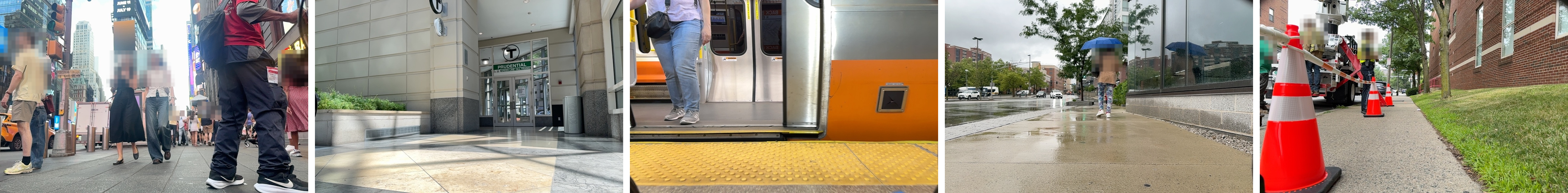}
    \caption{Representative scenes from the UNI dataset, illustrating the diversity of dataset, including crowded sidewalks, building entrances, transit interfaces, wet pavement, and construction-constrained walkways}
    \label{fig:dataset}
\end{figure*}

\section{UNI Overview}

In this section, we describe the UNI collection setup and the post-processing pipeline used to recover metric navigation trajectories from the recorded sensor streams. To preserve anonymity during peer review, we will publicly release the app source code, data-processing pipeline, trained model weights, and privacy-processed dataset after the review process concludes. Bill of materials (BOM) and other information can be found on website: \url{https://universal-nav.github.io/}.

\subsection{Rollator-Constrained Data Collection}

UNI uses a commercially available rollator as both a sensor carrier and physical feasibility filter. Unlike unconstrained human walking, the rollator cannot climb stairs, mount uncut curbs, or pass through gaps narrower than its footprint; consequently, the operator is naturally constrained toward routes compatible with wheeled mobility. The rollator requires approximately 30 minutes to assemble. A smartphone is attached using a standard phone mount, with no additional 3D-printed parts, custom-fabricated hardware, or platform-specific instrumentation required (Fig.~\ref{fig:teaser} right).

\subsection{SensorVault: Data Collection App}

For UNI data collection, we developed \textsc{SensorVault}, an iOS application that records calibrated multimodal sensor streams on LiDAR-equipped iPhones (Fig.~\ref{fig:teaser} right). The application supports any LiDAR-equipped iPhone; we used an iPhone~16~Pro for consistency across all collection sessions. The app records JPEG-compressed RGB images at $1280{\times}720$, temporally smoothed ARKit LiDAR depth and confidence maps at $256{\times}192$, and 6-DoF visual--inertial odometry (VIO) poses, all at 30\,Hz. It additionally records 100\,Hz inertial measurements, GPS, and audio. RGB, depth, and VIO poses are synchronized at the ARKit-frame level, with VIO poses expressed relative to a coordinate frame initialized at the start of each recording. Intrinsic camera parameters for the RGB images and depth maps, including focal lengths and principal-point coordinates, are stored with each recording. Fixed focus is used to maintain consistent camera calibration. All sensor streams share a monotonic clock and are stored in MCAP format, with low-confidence or out-of-range depth measurements marked invalid before storage.


Because device VIO can drift or fail over long recordings, we recover navigation trajectories independently using the depth-anchored reconstruction pipeline as described in Section~\ref{sec:post-processing}. Across recorded VIO chunks spanning 10.8\,h, 34.9\% produced motion inconsistent with plausible human walking. We therefore record VIO for reference but do not use it as the released trajectory source.

\textsc{SensorVault} provides a convenient implementation of UNI by recording synchronized, calibrated RGB, depth, inertial, GPS, and VIO streams in a single file, while LiDAR depth enables direct metric anchoring and quality checks during post-processing. The UNI paradigm itself is not tied to this specific application: its core supervision is RGB observations paired with metric trajectories, allowing the same collection principle to extend to other smartphones when paired with an appropriate metric trajectory-recovery method.

\subsection{Post-Processing the Collected Data}
\label{sec:post-processing}

To recover metric navigation trajectories, we process RGB recordings with Depth Anything 3 (DA3)~\cite{depthanything3} in 48-frame windows at 4\,Hz ($\sim$12\,s), with 12-frame (3\,s) overlap. DA3 jointly predicts per-frame depth $\hat{D}_i$ and camera poses $T_i=[R_i\mid t_i]$, where $R_i\in SO(3)$ and $t_i\in\mathbb{R}^3$ denote world-to-camera rotation and translation. Both predicted depth and camera translation are defined only up to a common scale. Let $\hat{C}_i=-R_i^\top t_i$ denote the predicted camera center. For window $w$,
\begin{equation}
D_i^{m}=s_w\,\hat{D}_i,\qquad
C_i^{m}=s_w\,\hat{C}_i,
\end{equation}
where $s_w$ is the unknown metric scale.

Consecutive windows are similarity-aligned over their overlapping frames and chained into a common relative coordinate frame. The recovered relative scale is applied to both camera centers and predicted depths, yielding $\tilde{C}_i$ and $\tilde{D}_i$ in a shared arbitrary unit. We reject inconsistent joins using the RMS disagreement between overlapping pose estimates.

We recover one absolute scale per chained segment using the smartphone's co-registered LiDAR depth $S_i$. Invalid, near-range, and range-saturated measurements are excluded. Let $\mathcal{F}$ denote frames with sufficient valid depth support and $M_i$ the valid pixels after resampling $S_i$ onto the predicted depth grid. We compute a robust per-frame scale ratio and take its median across the segment,
\begin{equation}
\rho_i=\operatorname*{median}_{p\in M_i}
\frac{S_i(p)}{\tilde{D}_i(p)},
\qquad
\hat{s}=\operatorname*{median}_{i\in\mathcal{F}}\rho_i .
\end{equation}

Segments without sufficient metric support are rejected rather than assigned a forced scale. This removes 8.3\% of candidate segments, which consist primarily of unusable or non-navigational recordings, such as camera occlusion or transport of the rollator. The filter therefore also serves as a quality check on the released navigation data. The resulting metric trajectory is $C_i^{m}=\hat{s}\,\tilde{C}_i$. We refer to this method as DA3+Anchor throughout the remainder of the paper.

For navigation training, future poses are expressed relative to the current observation and exported as metric trajectory targets paired with visual goals.

\section{The UNI Dataset}
\label{sec:uni_sidewalk}
The UNI dataset is an egocentric, rollator-constrained navigation corpus comprising 87 recording sessions collected by six operators in three cities across the United States (Boston, New York City and Worcester). Collection took place intermittently between April and August 2026, with an initial target of 10\,h to evaluate the collection pipeline and its utility for navigation learning. We currently operate two UNI systems to support parallel collection and continued dataset expansion.

The current corpus contains 10.8\,h of recordings, covering an estimated 37.2\,km of travel. Recordings span varied urban environments, weather conditions, times of day, transit by train, elevator, yielding to pedestrians, and waiting at crosswalks.

The dataset covers diverse surfaces across urban, indoor, and natural environments, including concrete sidewalks, brick and tactile paving, asphalt paths and roads, gravel, cobblestone, boardwalks, dirt, grass, sand, carpet, tile, and terrazzo, with examples of wet pavement. Representative scenes are shown in Fig.~\ref{fig:dataset}.

\subsection{Collection-Time Wheeled Feasibility}

UNI constrains route selection \emph{during collection}: operators follow paths traversable by the rollator, using curb cuts, ramps, lifts, and sufficiently wide passages. Throughout the 10.8\,h corpus, operators maintained this constraint rather than carrying the device over stairs or uncut curbs. When an intended route\textbf{} was inaccessible to the rollator, collection continued along an accessible alternative. For example, when a planned transit stop lacked accessible egress, the collector remained on the train and disembarked at the next accessible station.

The recordings capture repeated encounters with infrastructure relevant to wheeled mobility. Across the analyzed 10.8\,h corpus, we identify 169 crossing traversals, along with 49 lift rides across 19 sessions and 1.28\,km of tactile paving across 67 sessions. To audit adherence to the wheeled-feasible collection protocol, we flag candidate stair traversals using IMU stepping signatures and curb transitions using depth-based ground-plane fitting and step-height estimation. Video review of these candidates found no confirmed traversals of stairs or uncut curbs. These observations support the collection protocol's intended bias, while transfer to a particular robot additionally depends on its footprint, ground clearance, and allowable slope.

\subsection{Motion Distribution Relative to Robot-Native Data}

UNI also captures naturally occurring low-motion behavior. Because collection takes place in public pedestrian environments, demonstrations include waiting at crossings, yielding to pedestrians, and stop--resume transitions rather than only continuous forward motion.

We classify a window as near-stationary when its cumulative 2-D path length over five consecutive frame-to-frame steps is below 0.25\,m. Pooling eligible windows within each corpus, 8.14\% meet this criterion in UNI, compared with 3.10\% in EgoWalk and 0.25\% in SCAND (Table~\ref{tab:dataset_comparison}). Tartu/Milrem, in contrast, removes motion below 0.05\,m/s during preprocessing. UNI therefore preserves low-motion behavior that is less prevalent, or explicitly filtered, in several existing navigation corpora.

In Sec.~\ref{sec:experiments}, we test whether the two properties emphasized here---wheeled-feasible route selection and low-motion supervision---translate into useful learned navigation behavior.

\subsection{Model-Ready Export and Release}

We will release privacy-processed recordings in MCAP format, with faces and license plates blurred using EgoBlur~\cite{raina2023egoblur}. We will additionally provide a model-ready export of RGB observations and metric trajectories. The current snapshot comprises 1,087 trajectories and 117,512 frames sampled at approximately 4\,Hz, spanning varied environments, surface conditions, weather, and times of day. Data collection is ongoing, and the post-review release is expected to extend beyond this snapshot; all results reported here use the fixed dataset described in this paper.

\section{Experiments and Results}
\label{sec:experiments}
We engineered UNI and the UNI dataset to address four key research questions (RQs):
\begin{enumerate}
    \item \textbf{RQ1:} Can we recover reliable metric trajectories for navigation supervision without robot?
    \item \textbf{RQ2:} Does UNI improve trajectory prediction across navigation architectures?
    \item \textbf{RQ3:} Does retaining stationary demonstrations improve low-motion prediction?
    \item \textbf{RQ4:} Does the learned behavior transfer to robots?
\end{enumerate}
\begin{table}[t]
    \centering
    \caption{Metric trajectory recovery with depth inputs capped at 5\,m.
    \emph{Windows scored} gives reconstructed/total evaluation windows; metric
    ATE is computed after rigid SE(3) alignment without scale correction.}
    \label{tab:metric_recovery}
    \small
    \setlength{\tabcolsep}{3.5pt}
    \begin{tabular}{@{}llrrr@{}}
        \toprule
        Dataset & Method
        & \shortstack[r]{Windows\\scored}
        & \shortstack[r]{Scale err.\\(\%) $\downarrow$}
        & \shortstack[r]{Metric ATE\\(m) $\downarrow$} \\
        \midrule

        \multirow{5}{*}{TUM}
        & DA3 native       & 65/65 & 7.67 & 0.0455 \\
        & DA3 + anchor     & \textbf{65/65} & \textbf{3.14} & \textbf{0.0256} \\
        & VGGT + anchor    & 65/65 & 3.55 & 0.0286 \\
        & DROID + anchor   & 65/65 & 5.39 & 0.0450 \\
        \addlinespace[2pt]
        & ORB-SLAM3 RGB-D  & 64/65 & 2.08 & 0.0197 \\
        \midrule

        \multirow{5}{*}{KITTI}
        & DA3 native       & 82/82 & 22.92 & 2.483 \\
        & DA3 + anchor     & \textbf{80/82} & \textbf{6.75} & \textbf{0.720} \\
        & VGGT + anchor    & 80/82 & 6.99 & 0.785 \\
        & DROID + anchor   & 79/82 & 33.32 & 3.290 \\
        \addlinespace[2pt]
        & ORB-SLAM3 RGB-D  & 47/82 & \textit{5.01}$^{\dagger}$
                                   & \textit{0.018}$^{\dagger}$ \\
        \bottomrule
        \multicolumn{5}{@{}l@{}}{\footnotesize $^{\dagger}$Partial tracks:
        14\% frame completeness; not comparable.}
    \end{tabular}
\end{table}
\subsection{RQ1: Can we recover reliable metric trajectories for navigation supervision without robot?}

Because UNI recovers trajectory supervision without robot odometry, we first validate its accuracy against independent pose references. We evaluate on TUM RGB-D~\cite{tum}, which provides indoor motion-capture ground truth, and KITTI~\cite{kitti}, which tests outdoor reconstruction under faster vehicle motion using GPS/INS reference poses. Our navigation targets are future waypoints expressed relative to the current camera pose. A common rigid transformation of the reconstructed segment leaves these targets unchanged, whereas scale errors and local trajectory distortions directly affect supervision. We therefore evaluate short windows to assess metric accuracy over the local motion relevant to navigation training. Across 65 TUM and 82 KITTI windows, we report reconstruction coverage, median scale error, and metric absolute trajectory error (ATE) after rigid SE(3) alignment without scale correction. Depth inputs are capped at $5\,\mathrm{m}$.

On TUM, ORB-SLAM3~\cite{orbslam} RGB-D achieves the lowest reported errors, scoring 64/65 windows, while depth-anchored DA3 scores all 65 windows with a metric ATE of $0.0256\,\mathrm{m}$ (Table~\ref{tab:metric_recovery}). Anchoring reduces DA3's median scale error from 7.67\% to 3.14\%. On KITTI, it reduces scale error from 22.92\% to 6.75\% and metric ATE from $2.483$ to $0.720\,\mathrm{m}$, rejecting two windows with insufficient depth support. Under this short-range depth setting, ORB-SLAM3 produces only partial tracks in 47/82 windows, with 14\% frame completeness; its low reported ATE therefore does not represent full-window recovery. VGGT~\cite{vggt} with the same anchor achieves similar KITTI accuracy to DA3, whereas DROID~\cite{droid} retains substantial scale error, indicating that reconstruction quality remains important.

We next assess ORB-SLAM3 RGB-D on 30 UNI recordings to examine its suitability for corpus labeling and its agreement with our recovered trajectories. It initializes on all recordings but estimates poses for only 84.1\% of frames and produces complete trajectories for 10/30 recordings. Where tracking succeeds cleanly, its trajectories agree closely with ours, with rigidly aligned positional disagreement of 1.8--2.8\% of traveled distance. This agreement provides a consistency check on UNI, where independent ground truth is unavailable.


Together, these results support depth-anchored reconstruction for local metric navigation supervision. DA3 combines accurate short-window recovery with broader coverage than ORB-SLAM3 in the tested settings, while explicit quality rejection excludes segments lacking sufficient metric support.

\subsection{RQ2: Does UNI improve trajectory prediction across navigation architectures?}

We first evaluate whether UNI provides useful supervision across navigation architectures, then examine whether its benefits extend beyond the collection domain. For GNM~\cite{shah2023gnm}, ViNT~\cite{shah2023vint}, and NoMaD~\cite{sridharNoMaDGoalMasked2023}, we compare released checkpoints, UNI fine-tuning, and training from random initialization on UNI. Training runs for up to 20 epochs using AdamW with a learning rate of $5\times10^{-4}$, a cosine schedule with four warm-up epochs, and a batch size of 128.

Evaluation uses identical observation windows and goals from seven held-out outings disjoint from training, with five predicted waypoints per window. We report average displacement error (ADE) and $\mathrm{ADE}_{\mathrm{shape}}$, which fits one scalar per trajectory to account for differences in predicted motion magnitude. Errors are averaged within each trajectory and then equally across trajectories.

\begin{table}[htbp]
\centering
\caption{Held-out UNI trajectory prediction for ViNT checkpoints, training from scratch on UNI, and fine-tuning.}
\label{tab:comparison}
\resizebox{\columnwidth}{!}{%
\begin{tabular}{c ccc ccc}
\toprule
 & \multicolumn{3}{c}{ADE (m) $\downarrow$}
 & \multicolumn{3}{c}{$\mathrm{ADE}_{\mathrm{shape}}$ (m) $\downarrow$} \\
\cmidrule(lr){2-4}\cmidrule(lr){5-7}
Model & Released & Scratch & Fine-tuned
      & Released & Scratch & Fine-tuned \\
\midrule
GNM   & 0.318 & 0.251 & \textbf{0.239}
      & 0.204 & 0.188 & \textbf{0.161} \\
ViNT  & 0.322 & 0.297 & \textbf{0.245}
      & 0.211 & 0.220 & \textbf{0.176} \\
NoMaD & 0.321 & 0.307 & \textbf{0.265}
      & 0.217 & 0.216 & \textbf{0.186} \\
\bottomrule
\end{tabular}%
}
\end{table}
UNI fine-tuning reduces ADE by 17.4--24.8\% across the three architectures and improves scale-adjusted error (Table~\ref{tab:comparison}). Each model improves over its released checkpoint on all seven held-out outings. Fine-tuning also achieves lower ADE and scale-adjusted error than training from scratch within each architecture, although the ADE gap is small for GNM. These results establish the utility of UNI supervision on held-out data from its collection domain; we next examine whether the benefits extend to external corpora.

\begin{table}[htbp]
\centering
\caption{External-corpus evaluation of released and UNI-adapted
ViNT.}
\label{tab:cross_dataset}
\resizebox{\columnwidth}{!}{%
\begin{tabular}{llccc}
\toprule
Corpus & ViNT checkpoint & ADE (m) $\downarrow$
       & FDE (m) $\downarrow$ & Dir. ($^\circ$) $\downarrow$ \\
\midrule
\multirow{2}{*}{SACSoN}
 & Released & \textbf{0.242} & \textbf{0.403} & \textbf{8.10} \\
 & UNI fine-tuned & 0.261 & 0.443 & 10.51 \\
\midrule
\multirow{2}{*}{SCAND}
 & Released & 0.228 & 0.384 & 4.52 \\
 & UNI fine-tuned & \textbf{0.220} & \textbf{0.362} & \textbf{4.37} \\
\midrule
\multirow{2}{*}{CODa}
 & Released & \textbf{0.209} & \textbf{0.357} & 6.65 \\
 & UNI fine-tuned & 0.220 & 0.365 & \textbf{4.93} \\
\midrule
\multirow{2}{*}{SiT}
 & Released & 0.265 & 0.447 & 5.10 \\
 & UNI fine-tuned & \textbf{0.245} & \textbf{0.418} & \textbf{4.84} \\
\bottomrule
\end{tabular}%
}
\end{table}

We next evaluate released and UNI-fine-tuned ViNT on external corpora, using identical windows and a shared metric conversion within each corpus (Table~\ref{tab:cross_dataset}). SCAND and SACSoN contribute to released ViNT's pretraining mixture, so they test whether UNI adaptation retains performance on pretraining-source domains. The outcome is mixed: UNI fine-tuning improves all three metrics on SCAND, while released ViNT performs better on SACSoN. Adaptation therefore does not uniformly preserve or degrade performance across these domains.

On CODa~\cite{coda} and SiT~\cite{sit}, which are outside ViNT's pretraining mixture, the results likewise depend on the corpus and metric. On CODa, UNI fine-tuning reduces directional error by 25.9\%, while slightly increasing ADE and FDE. On SiT, it improves all three metrics, although evaluation covers only four trajectories. Thus, UNI's benefits extend beyond its collection domain, but do not constitute a consistent improvement in every measure of trajectory prediction.

\begin{table}[h]
\centering
\caption{Trajectory prediction on 39 CODa trajectories
after fine-tuning ViNT on UNI or EgoWalk.}
\label{tab:egowalk_comparison}
\small
\begin{tabular}{lccc}
\toprule
\addlinespace[3pt]
Training data & $\mathrm{ADE}_{\mathrm{shape}}$
              & $\mathrm{FDE}_{\mathrm{shape}}$
              & Heading error \\
              & (m) $\downarrow$ & (m) $\downarrow$
              & ($^\circ$) $\downarrow$ \\
\midrule
\addlinespace[3pt]
UNI     & 0.1355 & 0.2276 & \textbf{3.89} \\
EgoWalk & \textbf{0.1236} & \textbf{0.2161} & 6.12 \\
\hline
\end{tabular}
\end{table}

Finally, we ask how UNI compares with another source of human-collected supervision. Motivated by EgoWalk's analysis of turn-prediction errors~\cite[Sec.~5.2]{egowalk}, we fine-tune ViNT on approximately distance-matched UNI and EgoWalk subsets of 15.93\,km and 15.61\,km, respectively, and evaluate on CODa. UNI yields 36.4\% lower heading error, while EgoWalk has numerically lower scale-adjusted ADE and FDE; the latter differences are not statistically significant (Table~\ref{tab:egowalk_comparison}). This comparison supports a directional-prediction advantage in this setting, rather than overall superiority to EgoWalk.

Together, these results show that UNI improves held-out trajectory prediction across three navigation architectures. External evaluation with ViNT demonstrates benefits beyond the collection domain, including lower directional error on CODa, while the SACSoN regression shows that adaptation can trade off performance on previously learned domains.

\subsection{RQ3: Does retaining stationary demonstrations improve low-motion prediction?}

\begin{figure}[h]
    \centering
    \includegraphics[width=0.95\linewidth]{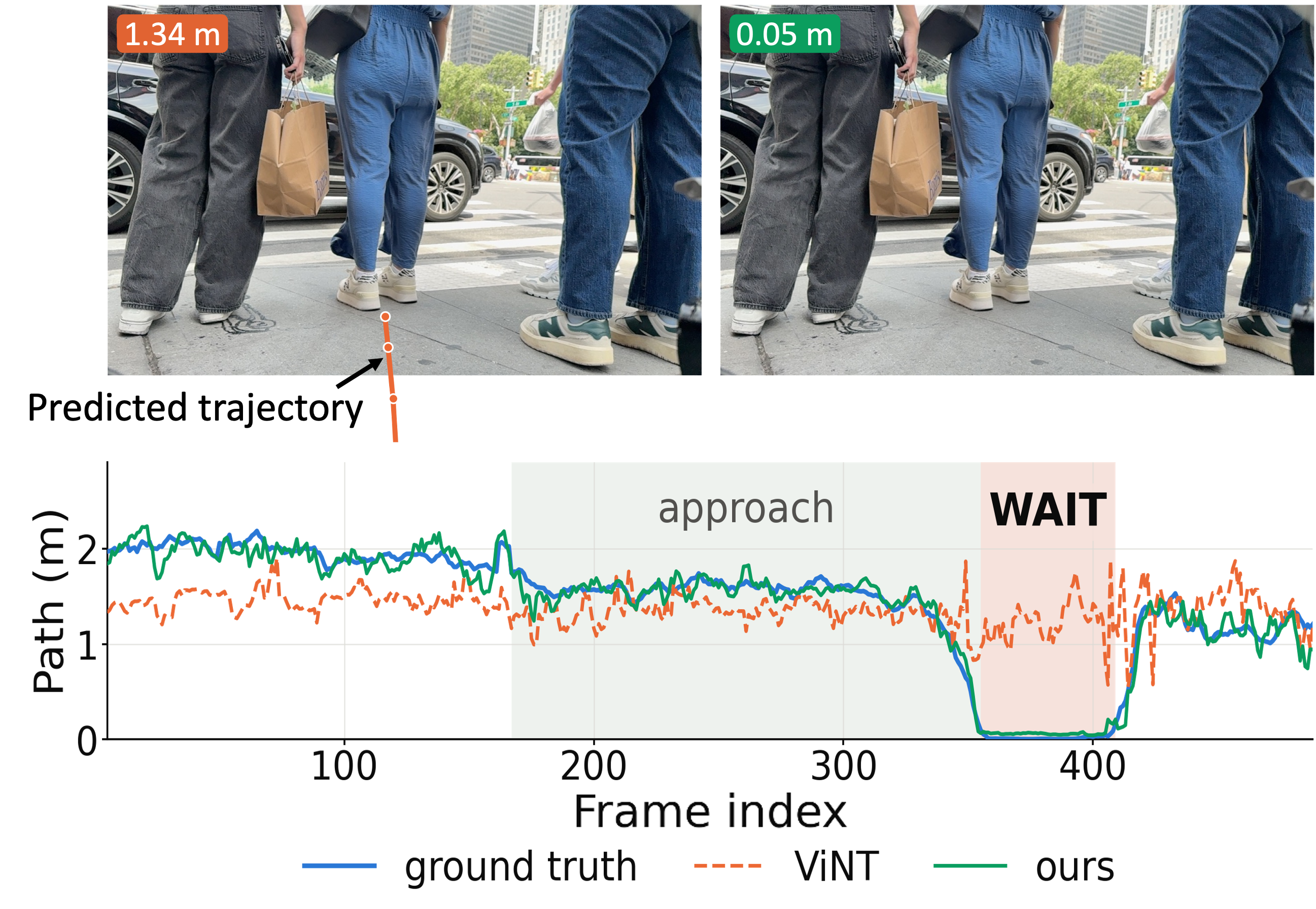}
     \caption{Held-out crosswalk pause with the goal image set to the current observation. ViNT~\cite{shah2023vint} fine-tuned on UNI predicts near-zero motion, while released ViNT continues to predict forward motion.}
    \label{fig:wait}
\end{figure}

\begin{table}[h]
    \centering
    \caption{Stationary supervision ablation on held-out UNI.
    $n$ denotes evaluation windows in each motion category.}
    \label{tab:rq3_stationary}
    \small
    \setlength{\tabcolsep}{4pt}
    \begin{tabular}{@{}lcc@{}}
        \toprule
        & \multicolumn{2}{c}{ADE (m) $\downarrow$} \\
        \cmidrule(lr){2-3}
        Policy & Moving & Stationary \\
        & ($n=1{,}825$) & ($n=505$) \\
        \midrule
        Released ViNT & 0.325 & 0.435 \\
        UNI fine-tuned & \textbf{0.286} & \textbf{0.149} \\
        \midrule
        Random removal & 0.288 & 0.200 \\
        Stationary removal & 0.295 & 0.336 \\
        \bottomrule
    \end{tabular}
\end{table}

Stationary behavior is less prevalent in several existing navigation datasets than in UNI (Table~\ref{tab:dataset_comparison}). We test whether retaining these demonstrations improves prediction during pauses without degrading prediction during motion. On held-out UNI windows, released ViNT has higher stationary than moving ADE ($0.435$ versus $0.325\,\mathrm{m}$). UNI fine-tuning reduces stationary ADE by 65.7\% while also improving moving ADE (Table~\ref{tab:rq3_stationary}). Figure~\ref{fig:wait} illustrates reduced motion prediction during a crosswalk pause when the goal image is set to the current observation.

To determine whether stationary examples specifically contribute to this improvement, we compare two training ablations: removing low-motion windows comprising 5.73\% of the training set, and removing the same number of randomly selected windows within trajectories. Both models are evaluated on the same held-out windows. Removing stationary examples increases stationary ADE by 68.0\% relative to random removal, while moving ADE changes little. Thus, retaining stationary demonstrations contributes to improved low-speed motion prediction on held-out UNI data beyond the benefit of training-set size alone.

\begin{figure}[h]
    \centering
    \includegraphics[width=\linewidth]{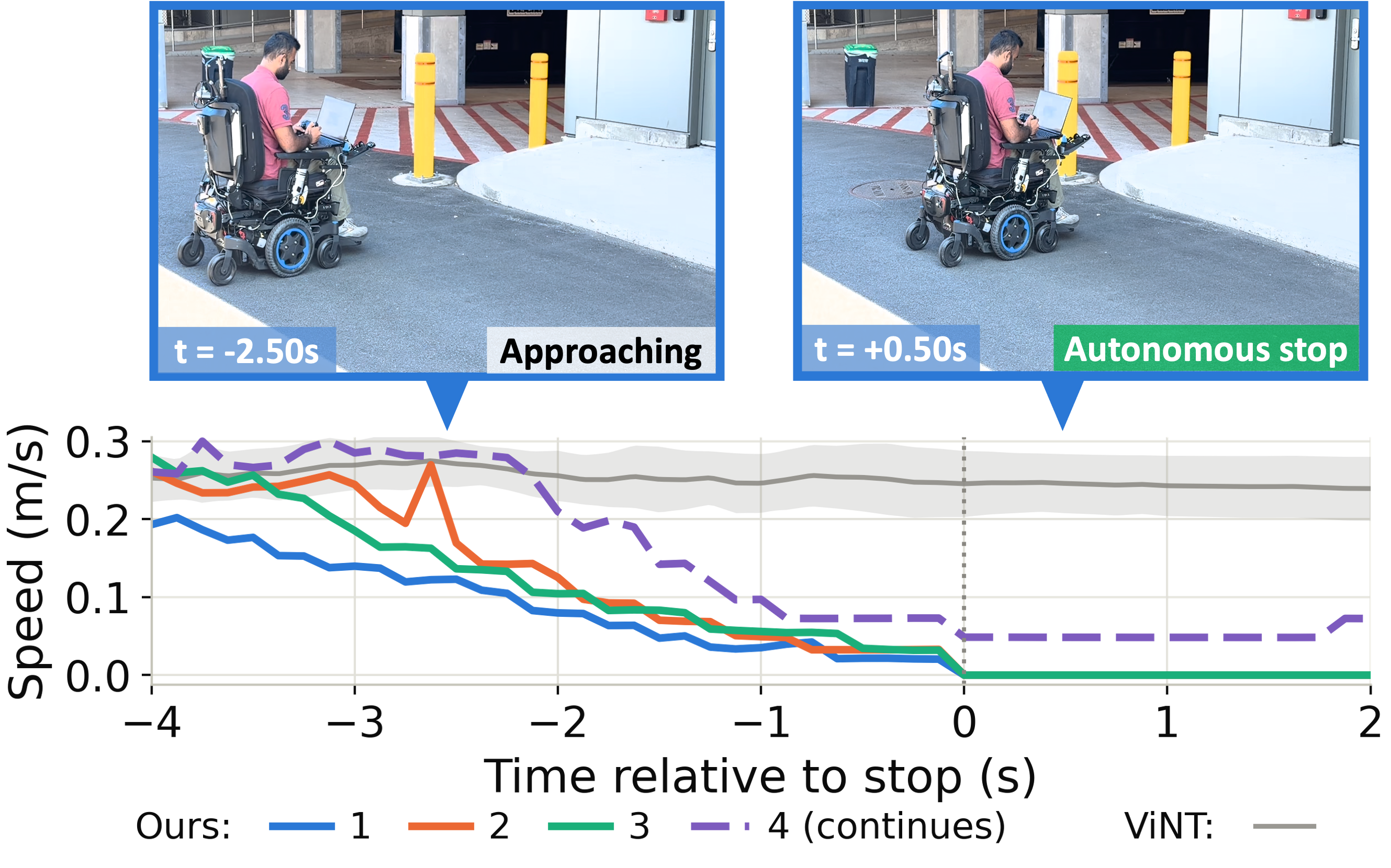}
    \caption{Closed-loop uncut-curb approaches. (Top) A representative autonomous stop with UNI-fine-tuned ViNT. (Bottom) Commanded speeds across four UNI trials and a released-ViNT baseline run. The wheelchair stops in three UNI trials and slows without stopping in the fourth.}
    \label{fig:wheelchair_curb}
\end{figure}
\subsection{RQ4: Does the learned behavior transfer to robots?}

We test whether fine-tuning on UNI demonstrations helps a powered wheelchair stop before uncut curbs and stairs while continuing through traversable curb cuts. We compare released and UNI-fine-tuned ViNT at four distinct locations per scenario near research campus. Each policy is tested once at each location, giving 12 trials per policy and 24 trials overall. Both policies receive RGB observation history and a goal image placed beyond the uncut curb, staircase, or traversable curb cut.

We deploy both policies on a Quickie Q500M powered wheelchair~\cite{quickie_q500m_manual}, with an iPhone mounted at $0.80\,\mathrm{m}$ as opposed to the $0.40\,\mathrm{m}$ collection height. \textsc{SensorVault} streams RGB observations to a Lenovo Legion 5i laptop (Intel Core i7-12700H, NVIDIA GeForce RTX 3070 Ti, $16\,\mathrm{GB}$ RAM) for online inference. Predicted waypoints are converted into linear and angular velocity commands and sent as ROS \texttt{Twist} messages to control software running on the wheelchair’s onboard Jetson Nano. Both policies share the same waypoint-to-velocity conversion, $0.30\,\mathrm{m/s}$ speed cap, and platform actuation threshold of $0.05\,\mathrm{m/s}$. A joystick multiplexer forwards policy commands only while the operator holds an enable button; releasing the button immediately disables autonomous command forwarding.

\begin{table}[h]
\centering
\caption{Success rate of closed-loop trials. Success is coming to rest before the non-traversable feature, or completing the curb-cut traversal without operator intervention.}
\label{tab:wheelchair}
\begin{tabular}{lcc}
\toprule
Scenario & Released ViNT & UNI fine-tuned \\
\midrule
Uncut curb (stop)   & 0/4 & \textbf{3/4} \\
Staircase (stop)    & 0/4 & \textbf{4/4} \\
Curb cut (traverse) & 4/4 & 4/4 \\
\bottomrule
\end{tabular}
\end{table}

UNI-fine-tuned ViNT stops before uncut curbs in 3/4 trials and staircases in 4/4 trials, compared with 0/4 for released ViNT in both scenarios (Table~\ref{tab:wheelchair}). Both policies complete all four curb-cut traversals. During traversal, mean commanded speed is $0.266\,\mathrm{m/s}$ for UNI-fine-tuned ViNT and $0.293\,\mathrm{m/s}$ for released ViNT. These averages exclude zero commands recorded after the wheelchair stops. Figure~\ref{fig:wheelchair_curb} shows the four uncut curb approaches: UNI reduces commanded speed and the wheelchair comes to rest in three trials, while slowing without stopping in the fourth. Released ViNT continues forward until operator intervention.

These results demonstrate transfer across embodiment and camera height: UNI adaptation improves stopping before the tested non-traversable features while preserving curb-cut traversal, with a 9\% reduction in mean commanded traversal speed. The remaining curb failure motivates further evaluation of traversability prediction and geometric reasoning.

\section{Discussion and Future Work}

UNI demonstrates that a simple physical rollator can provide useful navigation supervision without operating the target robot. Fine-tuning improves held-out prediction across three architectures, while comparison with EgoWalk shows a directional-prediction advantage on CODa. These findings support rollator-constrained collection as a complement to existing robot-native and human-collected data. Cross-dataset gains remain environment- and metric-dependent, motivating joint training with robot-native corpora to combine UNI's supervision with broader performance retention.

Retaining stationary demonstrations improves prediction during pauses, highlighting the value of collecting waiting and yielding alongside continuous motion. Predicting low motion, however, does not establish that a model understands why motion should be withheld or when it should resume. Future collection will include matched waiting and proceeding events with annotations of signal state, pedestrian interactions, and blocked paths. These examples will support evaluation of whether models respond to the scene conditions that govern stopping and resuming.

The wheelchair trials demonstrate transfer across embodiment and camera height, with improved stopping before the tested obstacles while preserving curb-cut traversal. The limited trial count and remaining curb failure motivate repeated evaluation across more locations, viewpoints, and approach conditions. The rollator biases route selection toward wheeled feasibility, but transfer still depends on the target platform's footprint and mobility constraints. Future work will investigate depth-informed traversability prediction and separate modeling of path geometry and motion timing to improve deployment reliability.

\section{Conclusion}

We introduced UNI, a robot-free navigation data collection interface using a commercial rollator and smartphone. UNI combines physically constrained human demonstrations with depth-anchored metric trajectory recovery, producing 37.2\,km of multimodal navigation data. Fine-tuning GNM, ViNT, and NoMaD reduces held-out trajectory error by 17.4--24.8\%, while the stationary-supervision ablation demonstrates the value of retaining low-motion examples. Closed-loop wheelchair experiments further demonstrate transfer from the collection interface to a different wheeled platform. Together, these results establish UNI as a practical, low-cost source of navigation supervision that complements robot-native data collection.

\section*{Acknowledgments}

The authors thank Adrian Patel, Asher Hunter, and Wisena Joseph for their assistance with data collection.

\bibliographystyle{IEEEtran}
\bibliography{references}

\end{document}